\documentclass[]{article}
\usepackage{cite}
\usepackage{amsmath}
\usepackage[utf8]{inputenc}
\usepackage{graphicx}
\usepackage{cleveref}
\usepackage{booktabs}
\usepackage{kotex}
\usepackage{multirow}
\usepackage{multicol}
\usepackage[super]{nth}
\usepackage{bm}

\begin{document}
\title{Batch Transformer: Look for Attention in Batch}

\author{Myung Beom Her, Jisu Jeong, Hojoon Song, and Ji-Hyeong Han\footnote{M.B. Her, J. Jeong, H. Song, and J.-H. Han are with the Department of Computer Science and Engineering, Seoul National University of Science and Technology, Seoul 01811, Republic of Korea (e-mail: gblader@seoultech.ac.kr; vaultroll@seoultech.ac.kr; ghwns0703@seoultech.ac.kr; jhhan@seoultech.ac.kr).\\This research was supported by the MSIT (Ministry of Science and ICT), Korea, under the ITRC (Information Technology Research Center) support program (IITP-2024-RS-2022-00156295) supervised by the IITP (Institute for Information \& Communications Technology Planning \& Evaluation). \\
Myung Beom Her and Jisu Jeong are co-first authors. Corresponding author: Ji-Hyeong Han.}}


\maketitle

\begin{abstract}
Facial expression recognition (FER) has received considerable attention in computer vision, with “in-the-wild” environments such as human-computer interaction. 
However, FER images contain uncertainties such as occlusion, low resolution, pose variation, illumination variation, and subjectivity, which includes some expressions that do not match the target label.
Consequently, little information is obtained from a noisy single image and it is not trusted. 
This could significantly degrade the performance of the FER task. 
To address this issue, we propose a batch transformer (BT), which consists of the proposed class batch attention (CBA) module, to prevent overfitting in noisy data and extract trustworthy information by training on features reflected from several images in a batch, rather than information from a single image. 
We also propose multi-level attention (MLA) to prevent overfitting the specific features by capturing correlations between each level. 
In this paper, we present a batch transformer network (BTN) that combines the above proposals. 
Experimental results on various FER benchmark datasets show that the proposed BTN consistently outperforms the state-of-
the-art in FER datasets. 
Representative results demonstrate the promise of the proposed BTN for FER.

\end{abstract}


\section{Introduction}

Facial expression recognition (FER) is an important computer vision task that classifies emotions based on human facial expressions. FER information is useful for computer vision applications such as social robots, human-computer interaction, and mental health monitoring. FER has been actively studied and shown promising results based on convolutional neural network (CNN) \cite{mt-arcres}, \cite{psr}, \cite{efficientface}, \cite{facebehaviornet}, \cite{adcorre}. However, existing methods have shown limitations in generalization ability. Therefore, researchers in FER are considering using a model based on the vision transformer (ViT) \cite{transfer}, \cite{apvit}, \cite{poster}, \cite{poster++}, \cite{arbex}. ViT, proposed for image classification, demonstrates excellent performance with the self-attention mechanism in the image processing tasks \cite{vit}. The attention mechanism, with the correlation between the patches, is the key reason for using ViT in FER \cite{transfer}. The most crucial aspect is to exclude human identity-information as much as possible and to focus on the information involved in facial expressions (e.g., eyes, nose, mouth, etc.) in FER \cite{pixel}. Attention allows for effective learning of emotional features by focusing on specific local areas in ViT.

However, ‘in-the-wild’ datasets collected from the internet still pose difficulties in emotion classification by the uncertainties of facial images. Uncertainties can degrade the quality of facial expressions in the image. According to Kiureghian et al. \cite{aleatory_epistemic}, uncertainties can be divided into aleatoric uncertainty, which is data uncertainty, and epistemic uncertainty, which is model uncertainty. Specifically, data uncertainty in FER arises from ambiguous facial expressions and the subjectivity of annotators, preventing the acquisition of reliable information from facial expression images \cite{scn}. Some of the uncertainties in FER frequently occur due to occlusion, low resolution, pose, illumination, and demographic variations. Occlusion obscures facial expressions by covering or revealing only part of the face. The low resolution of the face image makes it difficult to recognize the emotions. Additionally, pose, illumination, and demographic variations cause facial expressions to appear different depending on the direction, brightness, color of light, and even race, gender, and age.

\begin{figure}[pt]
    \centering
    \includegraphics[width=3.5in ]{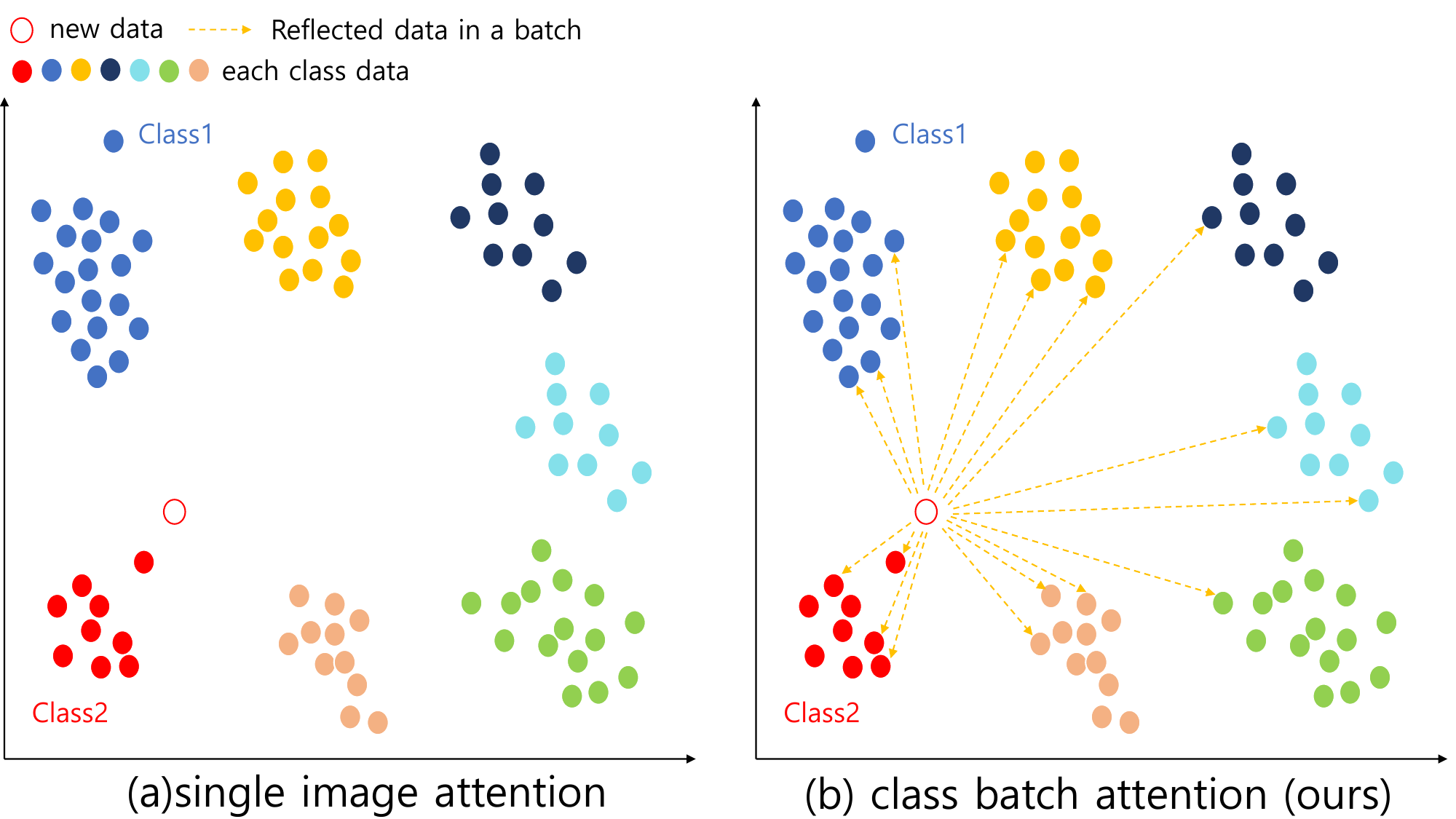}
    \caption{(a) shows multi-head self attention (MHSA), whose data cannot affect each other in training. (b) is class batch attention (CBA), whose data influence each other. It provides trustworthy information by reflecting class predictions of several images with similar features in a batch. We start from it to build the batch transformer.}
    \label{fig:CBA}
\end{figure}

Recently, attention mechanism models have been applied to suppress uncertainties and identify meaningful regions in facial expressions. PG-CNN \cite{pgcnn} develops a patch-gated CNN that integrates path-level attention to focus on occlusion, thus it distinguishes between occluded and unoccluded regions in FER. DSAN \cite{dsan} also adopts attention to address demographic variation. These models are focused on addressing uncertainties to improve performance. However, these previous attention models still exhibit low performance and struggle to reduce uncertainty.

To address the problem of FER, in this paper, we propose class batch attention (CBA). As shown in Fig. \ref{fig:CBA} (a), existing attention mechanisms such as multi-head self-attention (MHSA) train only one sample, and then extract features from a single image. Many datasets have a small percentage of uncertain data but may occasionally have data with unreliable information. Even a few of these noisy samples will significantly degrade the performance of FER. Moreover, the FER dataset contains a large amount of uncertain data, which leads to more overfitting problems. In contrast, as shown in Fig. \ref{fig:CBA} (b), our CBA can affect each data point by providing the predictions of emotion classes. It provides trustworthy information by reflecting the class prediction of images with similar feature maps. Additionally, it ensures that intra-class is closely related and inter-class is more distinct due to the property of using correlations between each sample.

This paper presents a batch transformer network (BTN) for reflecting information from other images. BTN consists of two bottom-up branches, multi-head cross attention (MHCA), vision transformer (ViT), multi-level attention (MLA), and batch transformer (BT) with class batch attention (CBA). Two bottom-up branches separately learn images and facial landmark features at different semantic levels. A facial landmark, a bundle of key points in a face image, provides important information from key areas such as the eyes, nose, and mouth for recognizing a person's facial expression. While high semantic level is spatially coarse and semantically strong, the low semantic level is spatially fine and semantically weak \cite{FPN}. To take advantage of these properties, we use MHCA with image features and landmark features at different semantic levels. However, an overfitting issue occurs in the attention layer of the lower semantic level due to the vanishing gradient, similar to what happens before ResNet \cite{resnet}.
To address this issue, we propose the MLA for capturing the landmark features at lower semantic level. The MLA helps neurons at the attention layer in lower semantic levels work actively by capturing the correlation with each semantic level, similar to ResNet. Through MLA, BT can obtain important information with correlations from low to high semantic levels. After embedding, MLA is forwarded to BT with class predictions from ViT to obtain the trustworthy information.

However, the mechanism of batch attention works the same way for inference. Thus, a model that trained by batch attention sometimes produces different inference results when the data contained in the batch changes.
Considering this critical problem, we propose the batch transformer (BT) for extracting the trustworthy information that reflects the features of several images and producing the same inference result when the data contained in the batch changes. 
The BT consists of CBA and the final classification is done from output of ViT.
ViT produces the same inference results by inferring before the batch attention mechanism.
CBA extracts features from information adjusted by class predictions by the property of CBA. 
BT combines the outputs of ViT and CBA, thus it extracts features from information that combines a sample with trustworthy information. 
The proposed three predictions, i.e., predictions from ViT, CBA, and BT, allow BTN to understand what emotions are included in the wild image.

In this paper, our contributions include the following:
\begin{itemize}
    \item We propose a batch transformer (BT) to alleviate overfitting from untrustworthy data by extracting information that reflects the features of several images in a batch and produce the same inference result.
    \item We present a class batch attention (CBA) to obtain the trustworthy information 
    by reflecting the prediction of the several images with similar features in a batch. Additionally, CBA ensures that intra-class is closely related while inter-class is more distinct due to the property of CBA. 
    \item We design a multi-level attention (MLA) to prevent overfitting from certain features by capturing the correlations between each semantic level. 
\end{itemize}

\begin{figure*}[h]
    \centering
    \includegraphics[width=\textwidth]{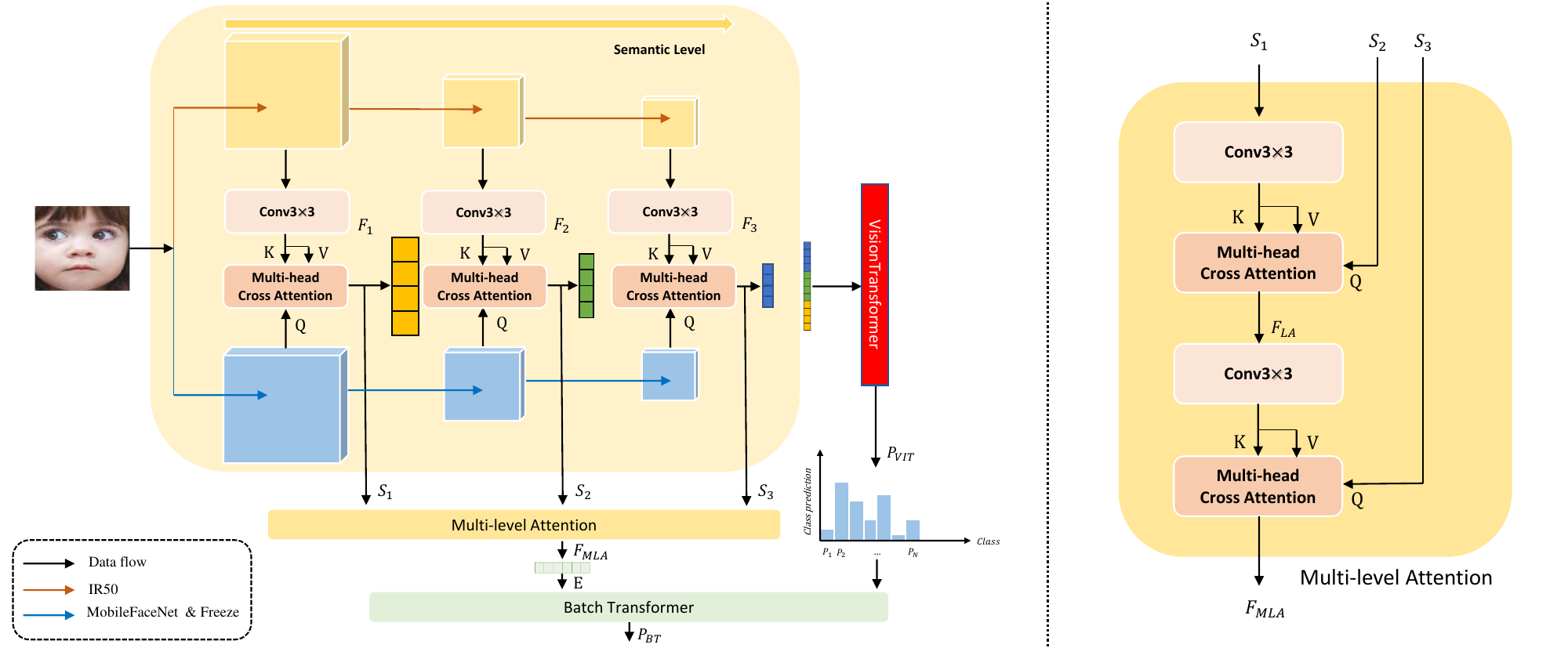}
    \caption{An overview of batch transformer network (BTN) and its sub-network architecture. The two pre-trained networks are employed for training BTN, one is IR50, which is pretrained with the MS-Celeb-1M for extracting image feature, and the other is frozen MobileFaceNet, which is pretrained with the MS-Celeb-1M for extracting landmark features. Each semantic level features of two backbones are forwarded to multi-head cross attention to capture attention for landmark features in image features. Captured features $S_l, l=1,2,3$ are forwarded to multi-level attention (MLA). After $F_{MLA}$ is embedding to E, it is forwarded to batch transformer (BT). N denotes the number of emotion labels.}
    \label{fig:BTN}
\end{figure*}

\section{Related Work}

\subsection {Facial Expression Recognition (FER)}

FER is a task in computer vision that recognizes and classifies emotions based on human facial expressions.  
Recently FER has highlighted the need for in-depth research for use in personalized services, psychotherapy, and education.
In response to this requirement, datasets in a laboratory environment such as CK+ \cite{ck+}, JAFFE \cite{jaffe}, and MMI \cite{mmi} were created based on Ekman's \cite{ekman} six basic human facial expressions (anger, surprise, disgust, happiness, fear, and sadness).

In the earliest study on FER, handcraft-based feature learning methods \cite{hog},\cite{lbp},\cite{sift} were presented. However, with the emergence of ‘in-the-wild’ datasets such as FER2013 \cite{fer2013}, AffectNet \cite{affectnet}, and RAF-DB \cite{rafdb}, which are closer to real-world conditions, existing methods have been replaced by deep learning such as convolutional neural network (CNN) and vision transformer (ViT) \cite{vit}.
‘In-the-wild’ datasets have remained a challenge due to uncertainties such as occlusion, pose variation, illumination variation, and subjectiveness of annotators.
To address this issue, Wang et al. \cite{scn} proposed self-cure network (SCN) to learn facial expression features and suppress uncertainty by relabeling samples with incorrect labels. Zhang et al. \cite{rul} presented relative uncertainty learning (RUL) to suppress uncertain samples by learning relative uncertainty through relativity with paired images. 
Zhang et al. \cite{eac} proposed erasing attention consistency (EAC), which learns by exploiting the flip attention consistency between the original and flipped images, but randomly erases images to avoid remembering noisy samples.

\subsection {Vision Transformers}

Since the great success of the transformer architecture in natural language tasks, there has naturally been research to apply to computer vision tasks as well. 
The vision transformer (ViT) \cite{vit} converted the image into patches and then used the encoder part of the transformer to determine the correlation between all patches, becoming one of the most studied architectures in the field of image processing.
The swin transformer \cite{swintransformer} was designed a hierarchical vision transformer to increase computational efficiency by utilizing cross-window connections with non-overlapping local self-attention. 
In the case of global context vision transformers \cite{gcvit}, long and short-range spatial interactions were more effectively modeled by utilizing global context self-attention in addition to existing local self-attention.

There have been many attempts to apply ViT in FER due to their powerful performance in various computer vision tasks. Aouayeb et al. \cite{vit+se} applied ViT by adding a squeeze-and-excitation (SE) block to ViT for FER work. TransFER \cite{transfer} designed the multi-head self-attention dropping (MSAD), which applied multi-head self-attention dropping to the transformer encoder for exploring the rich relationships between various local patches. More Recently, POSTER \cite{poster} proposed a two-stream pyramid cross-fusion transformer structure to explore the correlation between image features and landmark features. POSTER++ \cite{poster++} improved POSTER for a better balance between accuracy and computational complexity. POSTER++ used only POSTER's landmark-to-image branch and directly extracted multi-scale features from the backbone without using any additional pyramid architecture. The method improved FLOPs and reduced the number of parameters. 

However, the existing ViT-based models and the cross-attention models derived from the POSTER learned the relationships between patches within a single image. 
Compared to previous work, we aim to reflect the information of other samples for extracting trustworthy information.

\begin{figure*}[t]
\centering
\includegraphics[width=\textwidth]{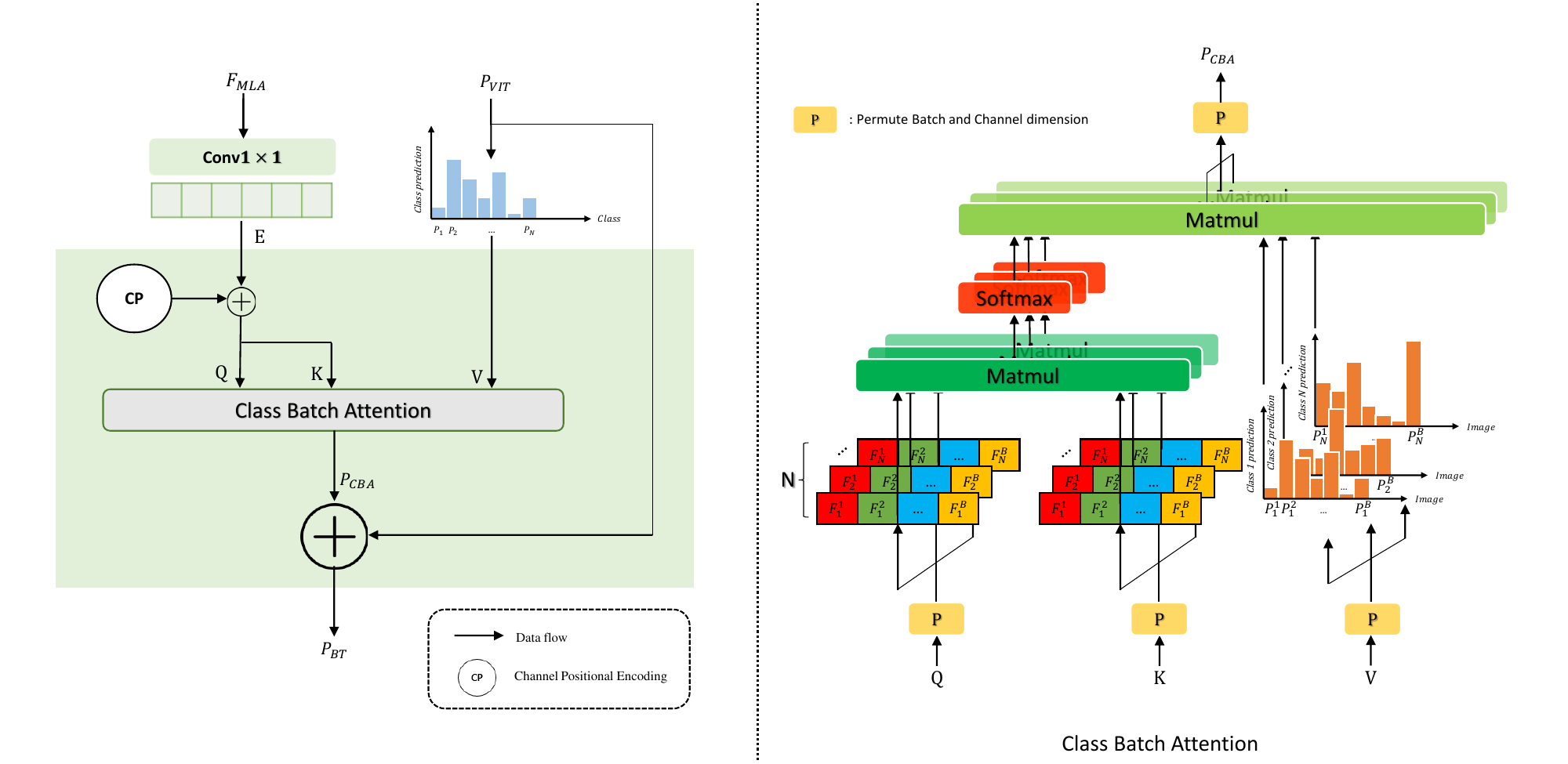}
\caption{Structure of the batch transformer. The feature map $F_{MLA}$ is embedded to $E$, using a convolution layer. Embedded features, $E$, are channel-positional encoded to obtain the same position per channel. After this, channel-positional encoded features and class prediction, $output_{VIT}$, are forwarded to class batch attention (CBA) for reflecting class prediction about several images with similar features. $P_{CBA}$ is added to $P_{VIT}$ for fusing features of a single image with features of several images. $F_{i}^{j}$, $P_{k}^{j}$ denote the $i^{\text{th}}$ channel feature map of the $j^{\text{th}}$ image in a batch and the $k^{\text{th}}$ class prediction of the $j^{\text{th}}$ image in the batch, respectively. B and N are the number of images in the batch and the number of class, respectively.}
\label{fig:BatchTransformer}
\end{figure*}

\section{Method}

In this section, we describe the batch transformer network (BTN) architecture with multi-level attention (MLA) and batch transformer (BT) with class batch attention (CBA). The BTN consists of two bottom-up branches for providing image features and landmark features at different semantic levels. 

\subsection{Batch Transformer Network} 
As shown in Fig. \ref{fig:BTN}, BTN consists of two bottom-up branches. One bottom-up branch generates image feature representations of different semantic levels, and the other bottom-up branch provides landmark features of different semantic levels. We employ IR50 \cite{arcface} as the backbone network of the first bottom-up branch to extract image information and MobileFaceNet \cite{mobilenet} as the backbone network of the second bottom-up branch to extract landmark information. Facial images are input to IR50 \cite{arcface} and MobileFaceNet \cite{mobilenet} to output feature maps of each semantic level. The semantic levels consist of three stages, such as low, middle, and high layers. The image feature maps in each semantic level are transformed to $F_l\in\Re^{C_l \times H_l \times W_l}, l=1,2,3$ using a convolution operation to fit the dimension of landmark features in each semantic level. The transformed image feature maps are separately forwarded into the multi-head cross-attention (MHCA) with landmark features to capture the attention of landmark features in image features and output $S_l\in\Re^{C_l \times H_l \times W_l}, l=1,2,3$. $S_l$ is embedded in the same channel and concatenated to the channel axis for fusing feature maps of each semantic level per channel, such as the swin transformer \cite{swintransformer}. Fused feature maps are forwarded to ViT.

Multi-level attention (MLA) consists of two MHCA and convolution operation, and $S_l$ is forwarded to the MLA. The low semantic information $S_1$ is forwarded to a convolution layer to fit $S_2$, and then MHCA with $S_2$ to capture correlations with low semantic level and mid semantic level. The captured features are forwarded to a convolution layer to fit $S_3$, and then MHCA with $S_3$ to capture the correlations between each semantic level. The MLA is formulated as:
\begin{equation}
	F_{MLA} = MHCA(S_3,conv(F_{LA}),conv(F_{LA}))
\end{equation}
\begin{equation}
    F_{LA} = MHCA(S_2, conv(S_1),conv(S_1))
\end{equation}
where $S_1$, $S_2$ and $S_3$ are low-level, mid-level, and high-level features, respectively, $conv(\cdot)$ is the convolutional neural network, and $MHCA(\cdot)$ is the multi-head cross-attention. 
Through the MLA, we can capture correlations between low and high semantic levels. Additionally, it provides an attention layer for capturing not only high-level features but also low-level features while the attention layers of the other models capture mostly high-level features. It helps alleviate overfitting by preventing dependence on high-level features. This is demonstrated in Fig. \ref{fig:MLAvis} by experiment.

The output of MLA is embedded to extremely small dimension to compress the information and leverage the memory efficiency and this is demonstrated in Fig. \ref{fig:Evis} by experiment. The Embedded output of MLA and the output of ViT are forwarded to batch transformer (BT) to obtain the reliable information and features by correcting the predictions. The final classification is performed from output of ViT to unchange the inference results when the data contained in the batch changes. Detailed process is described in the following Section \ref{BT}.

\subsection{Batch Transformer}\label{BT}

The FER images contain occlusion, low resolution, pose variation, illumination variation, and subjectiveness. Thus, little information can be obtained from a single image, which is difficult to trust. To address this issue, we propose a batch transformer (BT) to extract trustworthy features not from a single image but from information reflected in class predictions of several images with similar features. 
As shown in Fig. \ref{fig:BatchTransformer}, BT consists of the class batch attention (CBA).
When the number of classes is given by $N$, $F_{MLA}\in \Re^{C \times H \times W}$ is embedded to $E \in \Re^{N \times (H \times W )}$ using a convolution layer and view operation to extract the extremely important information. 
The embedded feature map $E$ is positional encoded per channel in order to have the same position per channel and then positional encoded $E$ and $P_{VIT}$ are forwarded to a CBA as query, key, and value. 
The output of CBA, i.e., $P_{CBA}$, and $P_{VIT}$ are added and it is the output of BT, i.e., $P_{BT}$.

The CBA consists of the permuted batch and channel dimension (P), the same query and key, which are the positional encoded embedded feature maps $E$, and the value, which is the class prediction $P_{VIT}$. 
Considering the batch dimension and the batch size given by $B$, the feature map of a single image $E\in \Re^{B\times N\times (H \times W )}$ is transformed to $F\in \Re^{1\times N\times B\times (H \times W)}$ using the P operation to bring the feature maps of images in a batch and the class prediction of the single image $P_{VIT}\in \Re^{B\times N}$ is transformed to $P\in \Re^{N \times B \times 1}$ using P operation to bring the class predictions of images in the batch. 
The transformed query $F$ and key $F$ are subjected to a dot product and softmax for each channel to get the similarity between the feature maps of images for each channel, and then we use the view operation and dot product with the transformed value $P$ to reflect the class predictions of images with similar feature maps for each channel. By using the P operation, we recover the original dimension. The CBA is formulated as:
\begin{align}
    &CBA(Q,K,V) = \text{P}[o_1, o_2, ..., o_N] \\
     &o_{n} = \theta(F_{n}F_{n}^T)P_{n}, n = 1, 2, ..., N \\
    &F = \text{P}(Q) = \text{P}(K),
    P = \text{P}(V)
\end{align}
where P$(\cdot)$ is the permuted batch and channel dimension, $N$ is the number of class, $B$ is the number of images in a batch, $F$ and $P$ denote the feature maps of the images in a batch and the class predictions of the images in the batch, respectively, $F_{n}$ and $P_{n}$ denote the $n^{\text{th}}$ channel feature maps of the images in the batch and the $n^{\text{th}}$ class predictions of the images in the batch, respectively, and $\theta(\cdot)$ is softmax. 
In CBA, the information is reliable and it prevents overfitting with unreliable information by reflecting the correlations with several images.
As a result, it provides intra-class relations that are closely related, while inter-class relations are more distinct. This is demonstrated in Fig. \ref{fig:tsne} by experiment.

$P_{CBA}\in \Re^{B \times N}$ is added to $P_{VIT}$ to obtain information combined a sample with trustworthy information. Finally, the BT is formulated as:
\begin{align}
    &P_{BT} = P_{CBA} + P_{VIT}\\ 
    &P_{CBA} = CBA(CP(E), CP(E), P_{VIT})\\
    &E = conv(F_{MLA})
\end{align}
where $F_{MLA}$ is output of MLA, which is feature map, $conv(\cdot)$ is a convolution layer, $CP(\cdot)$ is channel positional encoding, $CBA(\cdot)$ is class batch attention, $P_{VIT}$ and $P_{CBA}$ are outputs of ViT and CBA, respectively. 

By using multiple predictions as target of loss, we can learn the features on information about not only a single image but also several images. This helps prevents overfitting to noisy and untrustworthy information. This is demonstrated in Fig. \ref{fig:BTpredict} by experiment. Note that the $P_{VIT}$ is the final prediction, CBA and BT helps not in inference but in training. If the final prediction is done from a classifier after the CBA, the same mechanism will work for inference. Thus, if data in the batch change during inference, the prediction will change. Therefore, you must not perform inference after any batch attention mechanism. 
The final loss is formulated as:
\begin{equation}
	L = \lambda L_{VIT} + L_{BT} + L_{CBA}
\end{equation}
where $L_{VIT}$, $L_{BT}$, $L_{CBA}$ are losses applied to $P_{VIT}$, $P_{BT}$, and $P_{CBA}$, respectively, using cross-entropy (CE). Through these losses, we can extract features from information combined samples with trustworthy information.

\section{Experiment}

We analyze and compare the BTN with the leading convolutional neural network and vision transformer model on representative FER task. In addition, more experimental setups and ablation studies are presented.
\subsection{Datasets}
\textbf{RAF-DB}: RAF-DB \cite{rafdb} is a real-world database with more than 29,670 facial expression images, such as surprise, fear, disgust, and happiness, etc. These images contain natural facial expressions and are a significant challenge because of the subjectiveness of annotators, occlusion, pose variation, and low resolution. All samples on RAF-DB have been split into two subsets: a training set and a testing set.

\textbf{AffectNet}: AffectNet \cite{affectnet} is a large-scale database of facial expressions that contains two benchmark branches: AffectNet(7cls) and AffectNet(8cls). 
AffectNet(7cls) and AffectNet(8cls) contain 287,401 and 291,568 samples, respectively.  
The distribution of each class is extremely unbalanced. 
Specifically, in the training samples of AffectNet(7cls), happy contains 134,415 samples while disgust contains 3,803 samples only. 
Additionally, contempt contains 3,750 samples only in AffectNet(8cls).

\subsection{Implementation Details} 

\textbf{Preprocessing} All images are aligned and resized to 224 $\times$ 224 pixels and then interpolated to 112 $\times$ 112 pixels. The three stages of IR50 \cite{arcface} pre-trained by Ms-Celeb-1M \cite{msceleb} are used as an image feature extractor and the three stages of MobileFaceNet \cite{mobilenet} pre-trained by Ms-Celeb-1M \cite{msceleb} are used as a landmark feature extractor with frozen weights. 

\textbf{Training and Inference Details} We trained BTN with the Pytorch platform on an NVIDIA RTX 3090 GPU. We employed the SAM \cite{sam} optimizer with Adam \cite{adam} and ExponentialLR \cite{exponential} for training on all datasets. More specifically, on the RAF-DB dataset, we trained our model with an initial learning rate of 2e-5 and a batch size of 64. For the AffectNet(7cls) dataset, we trained our model with an initial learning rate of 0.8e-6 and a batch size of 144. For the AffectNet(8cls) dataset, we trained our model with an initial learning rate of 1e-6 and a batch size of 144. For data augmentation, we used random horizontal flipping as the default value in all datasets and random erasing with a scale=(0.02, 0.1) for RAF-DB, and p=1, scale=(0.02, 0.1) for AffectNet(7cls) and AffectNet(8cls). Finally, we used the ImbalancedDatasetSampler only in AffectNet to resolve the unbalanced class distribution problem.

\begin{table}[!t]
\caption{Comparison with state-of-the-art methods on RAF-DB. overall accuracy is implemented.}
\label{table_1}
\centering
\begin{tabular}{c c c}
\toprule
Method &Year &Accuracy \\
\midrule
SCN\cite{scn}  &	CVPR 2020 &	87.03 \\
PSR\cite{psr} & 	CVPR 2020 &	88.98 \\
LDL-ALSG\cite{ldl-alsg} &	CVPR 2020 &	85.53 \\
RAN\cite{ran} &	TIP 2020 &	86.90 \\
DACL\cite{dacl} &	WACV2020 &	87.78 \\
KTN\cite{ktn} &	TIP 2020 &	88.07 \\
DMUE\cite{dmue} &	CVPR 2021 &	89.42 \\
FDRL\cite{fdrl} &	CVPR 2021 &	89.47 \\
VTFF\cite{vtff} &	TAC 2021 &	88.14 \\
ARM\cite{arm} &	2021 &	90.42 \\
TransFER\cite{transfer} &	ICCV 2021 &	90.91 \\
DAN\cite{dan} &	2021 &	89.70 \\
EfficientFace\cite{efficientface} &	AAAI 2021 &	88.36 \\
MA-Net\cite{ma-net} &	TIP 2021 &	88.42 \\
Meta-Face2Exp\cite{meta-face2exp} &	CVPR 2022 &	88.54 \\
EAC\cite{eac} &	ECCV 2022 &	90.35 \\
APViT\cite{apvit} &	IEEE 2022 &	91.98 \\
POSTER\cite{poster} &	ICCV 2023 &	92.05 \\
POSTER++\cite{poster++} &	2023 &	92.21 \\
ARBEx\cite{arbex} &	2023 &	92.47 \\
BTN &	2024 &	\textbf{92.54} \\
\bottomrule
\end{tabular}
\end{table}

\begin{table}[!t]
\caption{Comparison with state-of-the-art methods on AffectNet(7cls). overall accuracy is implemented.}
\label{table_2}
\centering
\begin{tabular}{c c c}
\toprule
Method &Year &Accuracy \\
\midrule
PSR\cite{psr} &	CVPR 2020 &	63.77 \\
LDL-ALSG\cite{ldl-alsg} &	CVPR 2020 &	59.35 \\
DACL\cite{dacl} &	WACV 2020 &	65.20 \\
KTN\cite{ktn} &	TIP 2020 &	63.97 \\
DMUE\cite{dmue} &	CVPR 2021 &	63.11 \\
VTFF\cite{vtff} &	TAC 2021 &	61.85 \\
ARM\cite{arm} &	2021 &	65.20 \\
TransFER\cite{transfer} &	ICCV 2021 &	66.23 \\
DAN\cite{dan} &	2021 &	65.69 \\
EfficientFace\cite{efficientface} &	AAAI 2021 &	63.70 \\
MA-Net\cite{ma-net} &	TIP 2021 &	64.53 \\
Meta-Face2Exp\cite{meta-face2exp} &	CVPR 2022 &	64.23 \\
EAC\cite{eac} &	ECCV 2022 &	65.32 \\
POSTER\cite{poster} &	ICCV 2023 &	67.31 \\
POSTER++\cite{poster++} &	2023 &	67.49 \\
BTN & 2024 &	\textbf{67.60} \\
\bottomrule
\end{tabular}
\end{table}

\begin{table}[!htp]
\caption{Comparison with state-of-the-art methods on AffectNet(8cls). Overall accuracy is implemented.}
\label{table_3}
\centering
\begin{tabular}{c c c}
\toprule
Method &Year &Accuracy\\
\midrule
MT-ArcRes\cite{mt-arcres} & BMVC 2019 & 63.00 \\
SCN\cite{scn} &CVPR 2020&  60.23\\
PSR\cite{psr} &CVPR 2020&  60.68\\
ARM\cite{arm} & 2021 & 61.33\\
EfficientFace\cite{efficientface} & AAAI 2021&  60.23\\
MA-Net\cite{ma-net} &TIP 2021& 60.29\\
SL + SSL in-panting-pl\cite{sl+sslin-paranting-of} & 2022 & 61.72\\
Multi-task EfficientNet-B2\cite{multi-taskefficientnet-b2} & IEEE 2022 & 63.03 \\
DAN\cite{dan} &Biomimetics 2023 &62.09\\
POSTER\cite{poster} & ICCV 2023  & 63.34\\
POSTER++\cite{poster++} &	2023 &	63.77 \\
BTN & 2024 &	\textbf{64.29} \\
\bottomrule
\end{tabular}
\end{table}

\begin{table*}[!t]
\caption{Comparison with state-of-the-art methods on FER dataset. Class-wise accuracy and Mean Accuracy are implemented.}
\label{table_4}
\centering
\footnotesize
\hspace*{-2cm}
\begin{tabular}{c|c|cccccccc|cl}
\cline{1-11}
\multirow{2}{*}{Dataset} & \multirow{2}{*}{Method} & \multicolumn{8}{c|}{Accuracy of Emotions(\%)}                                                                          & \multirow{2}{*}{Mean Acc(\%)} &  \\
                         &                         & Neutral        & Happy & Sad            & Surprise       & Fear           & Disgust        & Anger          & Contempt &                               &  \\ \cline{1-11}
RAF-DB                   & MViT\cite{mvit}                   & 89.12          & 95.61 & 87.45          & 87.54          & 60.81          & 63.75          & 78.40          & -        & 80.38                         &  \\
RAF-DB                   & VTFF\cite{vtff}                    & 87.50          & 94.09 & 87.24          & 85.41          & 64.86          & 68.12          & 85.80          & -        & 81.20                         &  \\
RAF-DB                   & TransFER\cite{transfer}                & 90.15          & 95.95 & 88.70          & 89.06          & 68.92          & \textbf{79.37}          & \textbf{88.89}          & -        & 85.86                         &  \\
RAF-DB                   & POSTER++\cite{poster++}                & 92.06          & \textbf{97.22} & \textbf{92.89}          & 90.58          & 68.92          & 71.88          & 88.27          & -        & 85.97                         &  \\
RAF-DB                   & POSTER\cite{poster}                  & \textbf{92.35}          & 96.96 & 91.21          & 90.27          & 67.57          & 75.00          & \textbf{88.89}          & -        & 86.04                         &  \\
RAF-DB                   & APViT\cite{apvit}                   & 92.06          & 97.30 & 88.70          & \textbf{93.31}          & \textbf{72.97}          & 73.75          & 86.42          & -        & 86.36                         &  \\
RAF-DB                   & BTN                     & 92.21          & 97.05 & 92.26          & 91.49          & \textbf{72.97} & 76.25          & \textbf{88.89} & -        & \textbf{87.30}                &  \\ \cline{1-11}
AffectNet(7cls)          & APViT\cite{apvit}                  & 65.00          & 88.00 & 63.00          & 62.00          & 64.00          & 57.00          & \textbf{69.00}          & -        & 66.86                        &  \\
AffectNet(7cls)          & POSTER\cite{poster}                  & \textbf{67.20}          & 89.00 & 67.00          & 64.00          & \textbf{64.80}          & 56.00          & 62.60          & -        & 67.23                         &  \\
AffectNet(7cls)          & POSTER++\cite{poster++}                & 65.40          & \textbf{89.40} & \textbf{68.00}          & \textbf{66.00}          & 64.20          & 54.40          & 65.00          & -        & 67.45                         &  \\
AffectNet(7cls)          & BTN                     & 66.80          & 88.40 & 66.20          & 64.20          & 64.00          & \textbf{60.60} & 63.00          & -        & \textbf{67.60}                         &  \\ \cline{1-11}
AffectNet(8cls)          & POSTER\cite{poster}                  & 59.40          & \textbf{80.20} & 66.60          & 63.60          & 63.60          & \textbf{59.80}          & 58.80          & 54.71    & 63.34                         &  \\
AffectNet(8cls)          & POSTER++\cite{poster++}                & 60.60          & 76.40 & 66.80          & \textbf{65.60}          & 63.00          & 58.00          & 60.20          & \textbf{59.52}    & 63.76                         &  \\
AffectNet(8cls)          & BTN                     & \textbf{61.60} & 77.40 & \textbf{68.80} & \textbf{65.60} & \textbf{65.60} & 54.80          & \textbf{63.80} & 57.00    & \textbf{64.32}                & \\
\cline{1-11}
\end{tabular}
\hspace*{-2cm}
\end{table*}

\begin{table}[!t]
\caption{Results of ablation study in key components}
\label{table_5}
\centering
\begin{tabular}{l l r r}
\toprule
Method  &Accuracy\\
\midrule
Baseline (POSTER++)  & 92.21\\
MLA  & 90.97\\
BT & 91.98 \\
MLA + BT & \textbf{92.54} \\
\bottomrule
\end{tabular}
\end{table}

\begin{table}[!t]
\caption{Results of ablation study of $\lambda$}
\label{table_7}
\centering
\begin{tabular}{l l r r}
\toprule
$\lambda$  &Accuracy\\
\midrule
$\lambda$ = 1  &91.88 \\ 
$\lambda$ = 1.25 &91.98 \\ 
$\lambda$ = 1.5  &92.01\\
$\lambda$ = 1.75  &92.11 \\ 
$\lambda$ = 2.0  &\textbf{92.54} \\ 
$\lambda$ = 2.25  &92.24 \\ 
$\lambda$ = 2.5 &91.91 \\ 
$\lambda$ = 2.75  &92.24 \\ 
$\lambda$ = 3.0  &91.72 \\ 
\bottomrule
\end{tabular}
\end{table}

\begin{table}[!t]
\caption{Results of overall accuracy in several batchsize}
\label{table_8}
\centering
\begin{tabular}{c c c c c c c c}
\toprule
Batchsize  &16 &32 &64 &128 &144 &256\\
\midrule
RAF-DB&91.13 &91.66 &\textbf{92.54} &92.34 &92.43 &91.62\\
AffectNet(7cls) & 66.77 &67.23 &67.34 &67.31 &\textbf{67.60} &67.23\\
AffectNet(8cls) & 63.59 &63.74 &63.34 &63.77 &\textbf{64.29} & 63.82\\
\bottomrule
\end{tabular}
\end{table}

\begin{table}[!t]
\caption{Results of ablation study at Each Loss}
\label{table_6}
\centering
\begin{tabular}{l l r r}
\toprule
Loss Functions  &Accuracy\\
\midrule
$\lambda L_{VIT}$ & 92.14 \\
$\lambda L_{VIT} + L_{BT}$ &92.14 \\ 
$\lambda L_{VIT} + L_{CBA}$   &92.40 \\ 
$\lambda L_{VIT} + L_{BT} + L_{CBA} $  &\textbf{92.54} \\ 
\bottomrule
\end{tabular}
\end{table}

\subsection{Comparison with the State-of-the-Art Methods}

The proposed method is compared with the state-of-the-art on three datasets.
The evaluation metrics are overall accuracy and mean accuracy.
Overall accuracy is general accuracy which is calculated by averaging accuracy on all the test data.
Mean accuracy, which is calculated by averaging accuracy of each emotion label, is also needed to evaluate FER performance fairly, since FER dataset is greatly unbalanced on each emotion label. 
The best performance is marked in bold.

As shown in Table \ref{table_1}, we evaluate the proposed BTN on the RAF-DB dataset using overall accuracy.
BTN outperforms the state-of-the-art method, i.e., ARBEX \cite{arbex}, by 0.07\% point. 
Moreover, BTN has no data preprocessing tricks unlike ARBEX and the inference time is also faster than ARBEX due to detaching all proposed modules during inference time. 
As shown in Tables \ref{table_2} and \ref{table_3}, we evaluate the proposed BTN on the AffectNet 7cls and 8cls using overall accuracy.
BTN outperforms the state-of-the-art method, i.e., POSTER++ \cite{poster++} by 0.11\% point and 0.52\% point in the overall accuracy for 7 emotions and 8 emotions, respectively.
Table \ref{table_4} shows the accuracy for each emotion label and mean accuracy compared with the state-of-the-art methods.
The proposed BTN outperforms the state-of-the-art methods on all three dataset in mean accuracy.

\subsection{Ablation Studies}

In this section, we analyze the effectiveness of BTN, BT, and MLA, and then the performance differences according to the combination of loss functions and hyper-parameters are presented. All the experiments are conducted on the RAF-DB dataset.

\textbf{Evaluation of different components} We analyze the accuracy of baseline model, i.e., POSTER++, and the proposed method with only MLA, only BT, and both of MLA and BT as shown in Table \ref{table_5}. 
The results demonstrate that the proposed modules are effective for FER. 
In the second two rows, optimizing components individually degrade the performance of BTN. 
In contrast, the combined MLA with BT strategy succeeded by solving problems of FER in the 4th row. 
The results imply that BT needs MLA for extracting diverse level information.

\textbf{Evaluation of $\bm{\lambda}$} 
Table \ref{table_7} shows the performance 
when $\lambda$ ranges from 1.0 to 3.0. As $\lambda$ increases per 0.25, we can see a trend of improved performance. We acquire the
best performances at $\lambda =$  2.0 and then
the accuracy decreases rapidly from 2.0 to 3.0. This result proves the effect of 
 $L_{VIT}$ in BTN and implies that the balance  between coefficient of $L_{VIT}$, $L_{BT}$, and $L_{CBA}$  should be adjusted appropriately.

\textbf{Evaluation of batch size} 
Table \ref{table_8} shows the performance at FER datasets
when batch size ranges from 16 to 256. As batch size increases per twice, exceptionally employed 144, we can see a trend of improved performance. We acquire the best performances at batchsize = 64, 144 and then
the accuracy decreases rapidly in RAF-DB, AffectNet datasets, separately. This result proves too little and much data degrade the performance of BT.

\begin{figure*}[htp]
\centering
\includegraphics[width=\textwidth]{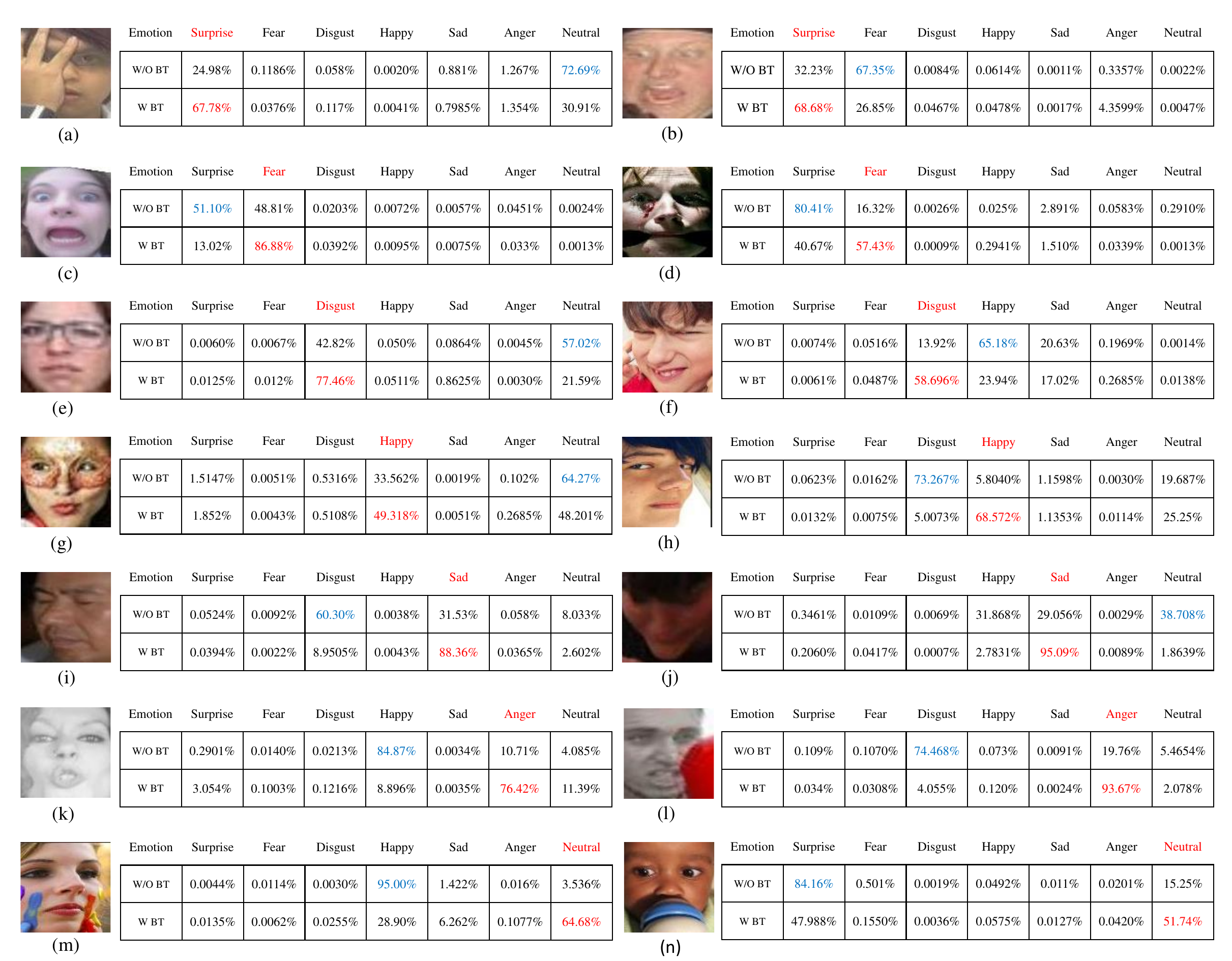}
\caption{Predicted probability distribution results of with batch transformer and without batch transformer on RAF-DB, especially hard cases such as occlusion, low resolution, pose variation, and illumination variation. Ground truth labels and correct predictions are marked in red.}
\label{fig:BTpredict}
\end{figure*}

\textbf{Evaluation of batch transformer} 
Fig. \ref{fig:BTpredict} shows the predicted probability for comparing with BT and without BT about occlusion, low resolution, pose variation, and illumination variation in RAF-DB. BT predicts labels accurately in (a),(c),(d),(g),(m), and (n),  
which contain occlusions, (b),(e), and (j), which contain low resolutions, (f),(h),(i), and (m), which contain pose variation, (d),(k), and (l), which contain illumination variation while without BT predicts labels incorrectly. It demonstrates that BT is robust on ``in-the-wild'' dataset by training trustworthy information.

\textbf{Evaluation of different loss functions}
Table \ref{table_6} verifies that the proposed losses are effective for FER. In the first three rows, optimizing components individually degrades the performance of BTN. In contrast, learning using a combination of losses succeeded by solving problems of FER in the 4th row. The results imply that $L_{VIT}$ needs $L_{BT}$ and $L_{CBA}$ for learning information of several images.

\subsection{Visualization}

\begin{figure}[!t]
\centering
\includegraphics[width=3.5in]{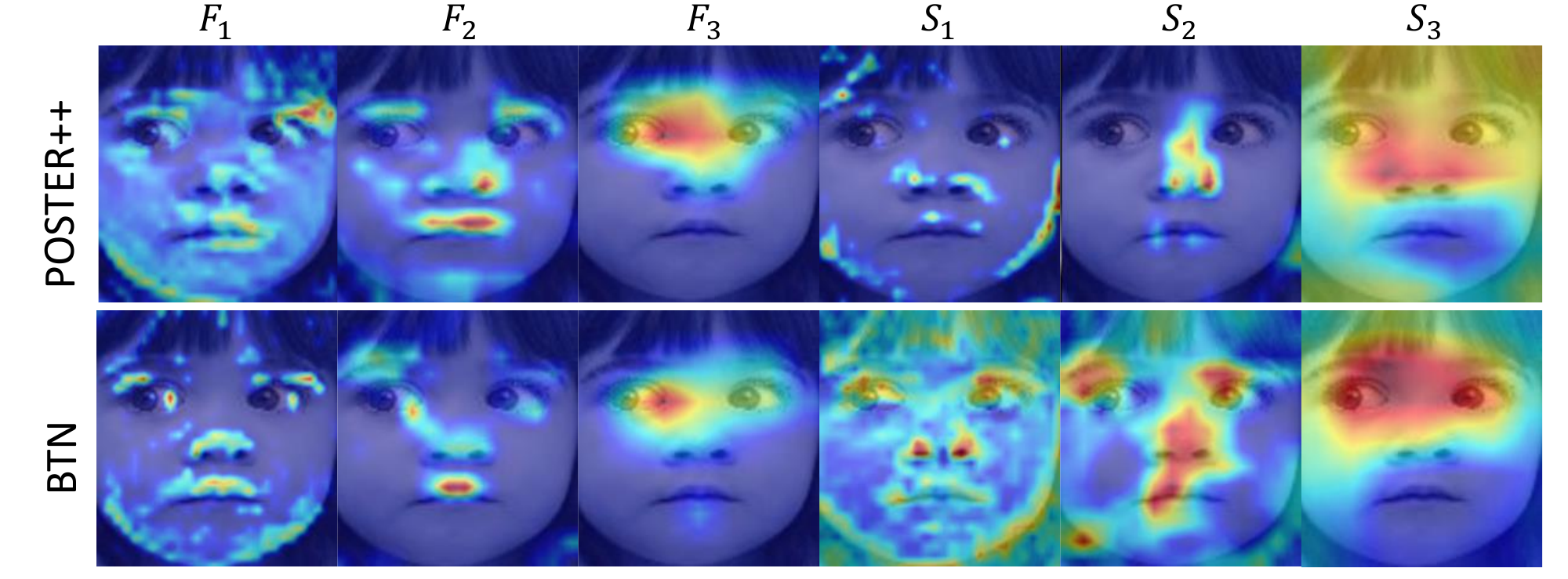}
\caption{Visualization of the activation map generated by Score-CAM for comparing the proposed BTN with POSTER++ in each semantic level.}
\label{fig:MLAvis}
\end{figure}

To demonstrate that the proposed BTN, BT, and MLA work as intended, we visualize the learned feature map. 
As shown in Fig. \ref{fig:MLAvis}, we visualize activation maps generated by Score-CAM \cite{scorecam} for the proposed MLA. We extract activation maps for the convolution layers and the MHCA at each semantic level. BTN and POSTER++ are similar in the feature extraction layers $F_1$, $F_2$, and $F_3$ while they are extremely different in the attention layers $S_1$, $S_2$, and $S_3$. The BTN captures many features from low to high level in the attention layer by employing MLA while POSTER++ cannot capture features in the low and mid levels. It demonstrates that the proposed MLA prevents overfitting by capturing features at each semantic level.

\begin{figure}[pt]
\centering
\includegraphics[width=3.5in]{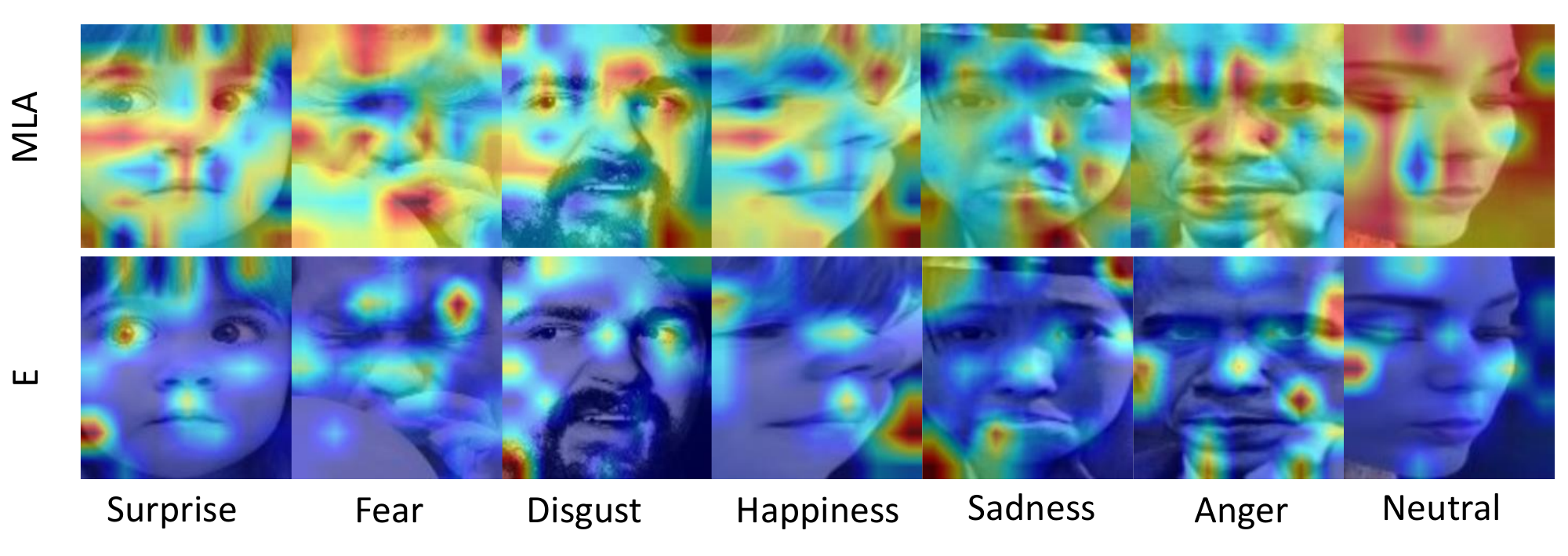}
\caption{Visualization of the activation map generated by Score-CAM for MLA and embedding layer of the proposed BTN.}
\label{fig:Evis}
\end{figure}

To verify that the embedding layer in BT works as intended, we visualize the learned feature map. As shown in Fig. \ref{fig:Evis}, we visualize activation maps generated by Score-CAM for the proposed MLA and embedding layer $E$ for different labels. The embedded feature map $E$ has activation maps in important parts while MLA has many activation maps broadly. It demonstrates that $E$ provides important information by decreasing the channel.

\begin{figure}[!t]
\centering
\includegraphics[width=3.5in]{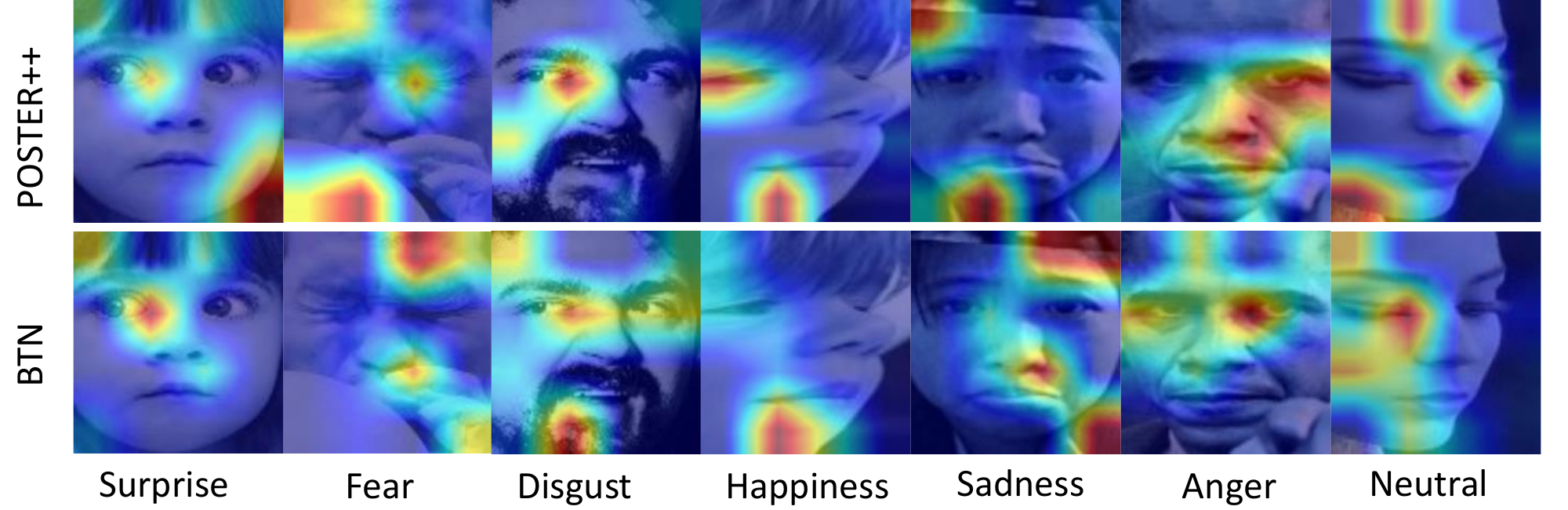}
\caption{Visualization of the activation map generated by Score-CAM for comparing the proposed BTN with POSTER++ in inference layer.}
\label{fig:BTNvis}
\end{figure}

To prove that the proposed BTN works as intended, we visualize the learned feature map. As shown in Fig. \ref{fig:BTNvis}, we visualize activation maps generated by Score-CAM for the proposed BTN and the baseline POSTER++ in the inference layer. BTN has a wider range of attention map than POSTER++. It proves that BTN is robust to noisy data by preventing dependence on specific features.

\begin{figure}[!t]
\centering
\includegraphics[width=3.5in]{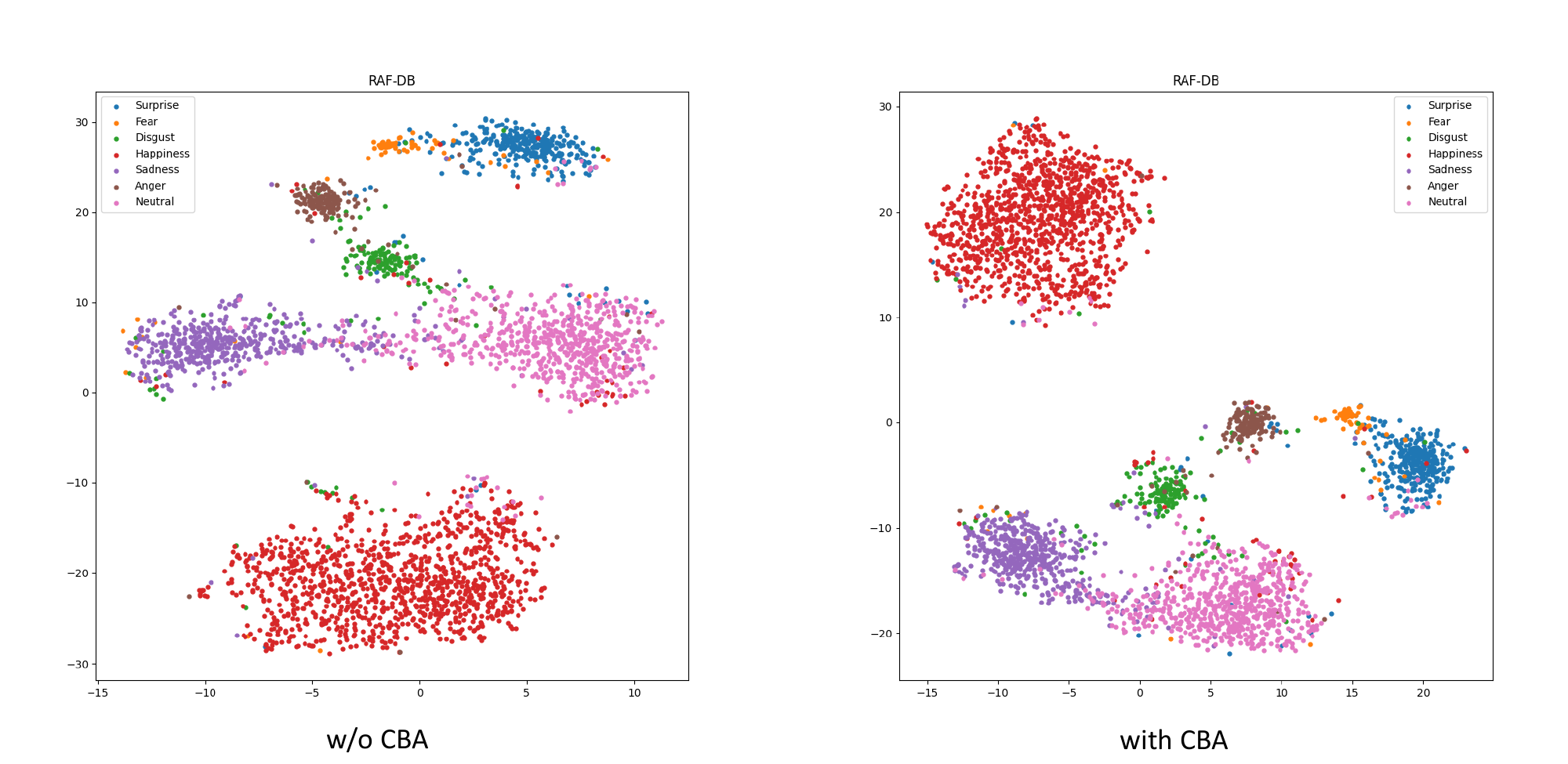}
\caption{t-SNE visualization of feature distribution about without CBA and with CBA.}
\label{fig:tsne}
\end{figure}

To verify that the proposed CBA works as intended, we utilize t-SNE \cite{tsne} to visualize feature distribution on RAF-DB as shown in Fig. \ref{fig:tsne}. We extract $output_{VIT}$ and implement t-SNE.
In the proposed CBA, it is clearly observed that the feature distribution of labels is clustered compared to that without CBA. Moreover, the boundaries of the feature distributions between different classes are distinguished in the model with CBA. It demonstrates that the proposed CBA ensures that the same labels are trained to have similar features. Therefore, it can help make intra-class features more closely related while inter-class features are more distinct.

\section{Conclusion}


FER has been a challenging task because these data contain uncertainty such as occlusion, low resolution, pose variation, illumination variation, and subjective, which include some expressions that do not match the target label.
However, existing methods have directly learned to extract discriminative features from a single image and cannot solve above mentioned problems. In this paper, we proposed the BTN with BT and MLA. The proposed BT prevented overfitting in noisy data and obtained trustworthy information by training on information reflected in the features of several images. The proposed MLA prevented overfitting to specific features by capturing correlations between each level. The evaluation results have shown that the proposed method exceeded state-of-the-art methods on FER datasets. Although BTN was an initial study about batch attention mechanism, its performance demonstrated that it has potential. Therefore, it is necessary to explore other batch attention mechanisms in future studies.

\bibliographystyle{ieeetr}
\bibliography{ref}

\end{document}